\documentclass[10pt,twocolumn,letterpaper]{article}

\usepackage{cvpr} % uncomment this for the final submission
\usepackage{times}
\usepackage{epsfig}
\usepackage{graphicx}
\usepackage{amsmath}
\usepackage{amssymb}
\usepackage{multirow}
\usepackage{makecell}
\begin{document}

%%%%%%%%% TITLE
\title{A Comprehensive Evaluation Framework for the Study of the Effects of Facial Filters on Face Recognition Accuracy}

\author{Kagan Ozturk\textsuperscript{*}, Louisa Conwill\textsuperscript{*}, Jacob Gutierrez, Kevin Bowyer, and Walter J. Scheirer\\
University of Notre Dame\\
Notre Dame, Indiana, USA\\
{\tt\small \{kztrk, lconwill\}@nd.edu, jacobgutier7@gmail.com, \{kwb, wscheire\}@nd.edu}
}
\maketitle

\def\thefootnote{*}\footnotetext{Equal contribution.}
\def\thefootnote{}\footnotetext{Kagan Ozturk is  supported by the Ministry of National Education of Türkiye.}

\thispagestyle{empty}

%%%%%%%%% ABSTRACT
\begin{abstract}
Facial filters are now commonplace for social media users around the world. Previous work has demonstrated that facial filters can negatively impact automated face recognition performance. However, these studies focus on small numbers of hand-picked filters in particular styles. In order to more effectively incorporate the wide ranges of filters present on various social media applications, we introduce a framework that allows for larger-scale study of the impact of facial filters on automated recognition. This framework includes a controlled dataset of face images, a principled filter selection process that selects a representative range of filters for experimentation, and a set of experiments to evaluate the filters' impact on recognition. We demonstrate our framework with a case study of filters from the American applications Instagram and Snapchat and the Chinese applications Meitu and Pitu to uncover cross-cultural differences. Finally, we show how the filtering effect in a face embedding space can easily be detected and restored to improve face recognition performance.
\end{abstract}

%%%%%%%%%%%%%%%%%%%%%%%%%%%%%%%%%%%%%%%%%%%%%%%%%%%%%%%%%%%%%%%%%%%%%%%%%%%%%%%%
%%%%%%%%% BODY TEXT
\section{Introduction}
\label{sec:intro}

\begin{figure}
    \centering
    \includegraphics[width=0.35\textwidth]{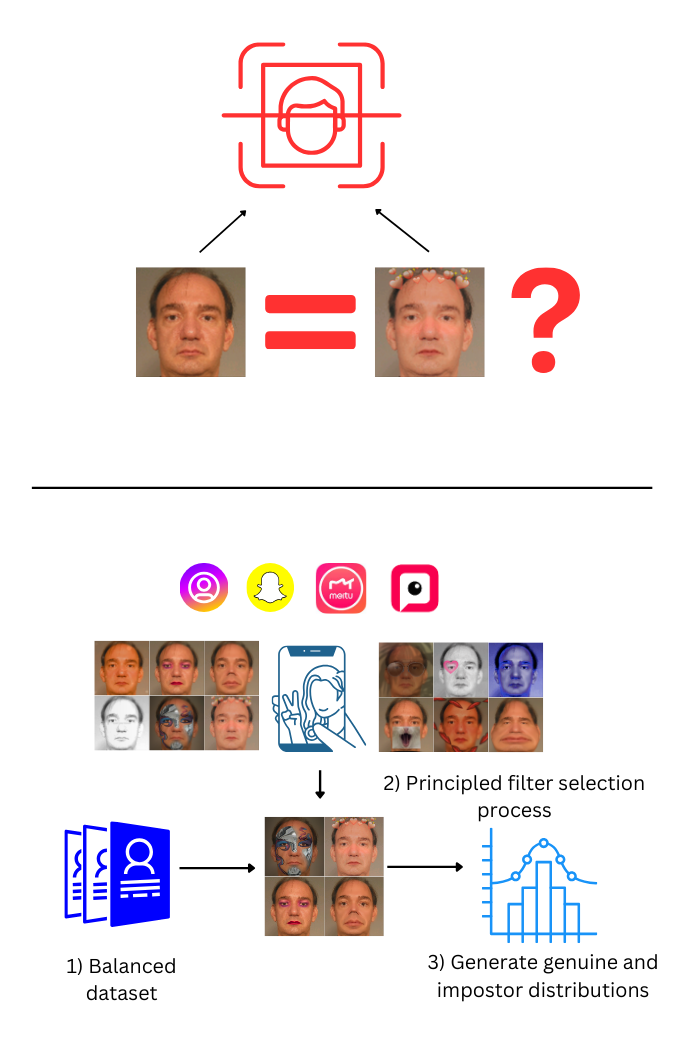}
    \caption{How do filtered faces affect automated recognition? We propose a framework that enables the study of this question on a per-filtering app basis. Our framework includes a controlled and gender-balanced database of images, a filter selection method that ensures a representative set of filters are selected from a particular platform, and a set of experiments consistent with accepted practice in the biometrics literature.}
    \label{fig:framework}
    \vspace{-5pt}
\end{figure}

Augmented Reality (AR) facial filters allow us to creatively re-imagine how our faces look by overlaying virtual content onto the image stream from a smartphone camera. Image-based content has become an essential and captivating part of social media sites \cite{choi2018instagram, kostyk2021perfect} and there are a number of tools that enable users to make their images more visually appealing before posting \cite{vendemia2018effects}. Popular modifications include changing the color tones of the image, changing the appearance of skin texture or facial structure, overlaying makeup to a face \cite{Shein_2021}, or transforming one's face into the likeness of Darth Vader or a puppy. Filtered photos now drive a significant amount of engagement on social media \cite{lavrence2020look}.

Filtered face images are ubiquitous on social media, and can reduce the accuracy of facial recognition systems \cite{ferrara2016effects, rathgeb2019impact}. This can lead to negative societal effects, especially with relation to police investigations. Police are increasingly using social media to aid investigations \cite{Kelly_2012}, feeding images from social media to face recognition algorithms~\cite{Barrett_2021}. This can help for especially heinous crimes like human trafficking, for which police use face recognition to help find traffickers and victims \cite{Joshua_Lee_2023, Foy_2021}. Face recognition's decreased accuracy with filtered images could make it harder to find missing persons \cite{Carneiro_2019}. Because of this, it is important to further our understanding of how filters impact face recognition systems.

There is previous work  studying the  effect of social media facial filters on face recognition \cite{botezatu2022fun, hedman2022effect, mirabet2022impact, riccio2022openfilter, tiwari2023frll, mirabet2024facial}. However, each previous work only considers a few filters. To our knowledge there are no studies that consider more than 10 different filters. These filters are often hand-picked by the authors and fit particular criteria, for example filters that occlude the face or beautification filters. This selective approach raises concerns about the generalizability of the findings, as the vast array of filters available on social media platforms remains largely unexamined. Given that images from these platforms are increasingly used in critical facial recognition tasks, such as identifying missing persons, it is crucial to understand how the full scope of filters on a particular platform may affect recognition accuracy. The literature currently lacks a framework for studying a large and representative sample of filters from particular facial filtering applications to evaluate how images from that particular platform may impact recognition. Thus, the main contribution of this paper is a framework for a more comprehensive study of facial filter effects on facial recognition, with the following components:
\begin{itemize}
    \item A controlled, gender-balanced image set.
    \item A filter selection process that ensures a large-scale, diverse, and representative set of filters are chosen from a particular filtering app.
    \item A protocol for evaluating the selected filters' impact on recognition.
\end{itemize}

In this paper, we also demonstrate how our framework can be used through a case study of filters from two American apps Instagram and Snapchat and two Chinese apps Meitu and Pitu. Through our case study we make the following additional contributions:
\begin{itemize}
    \item Analysis of the impact of the largest number of facial filters (125 total filters across 4 social media platforms) on recognition.
    \item The first cross-cultural comparison of filters from Western and Asian apps to our knowledge, and a discussion of the different social implications of facial filter usage in these regions in light of our findings.
    \item A filter mitigation approach for enhancing recognition performance.
\end{itemize}

%-------------------------------------------------------------------------
\section{Related Work}
% need to cite this paper too https://ieeexplore.ieee.org/stamp/stamp.jsp?tp=&arnumber=10623751

Botezatu \etal \cite{botezatu2022fun} looked at the impact of augmented reality selfie filters on face recognition, and focused on facial occlusions caused by filters. They used ten filters from the selfie editing apps Sweet Face Camera, B612, Snow, YouCam Fun, and Bloom Camera. They applied the filters to the FRGCv2 dataset using an Android emulator. 
To evaluate the recognition performance, they assess the False Non-Match Rate (FNMR) at three different False Match Rates (FMR) for both ArcFace and a COTS facial recognition system. 
They find that facial images with high selfie filter coverage and facial images with occluded mouths and noses negatively impact facial recognition the most. 

Hedman \etal \cite{hedman2022effect} considered the effects of 9 filters from Instagram that change the color tones of the image and 4 filters that occlude either the eyes or the nose. They applied the filters to the Labeled Faces in the Wild and CelebA datasets. They used a ResNet34, SqueezeNet, and ResNet50. They considered identification using various distance measures plus SVM and XGBoost to find the closest subject in the database. They considered both closed and open set experiments for identification, and considered the FNIR and GAR in the closed set experiments and FPIR and FNIR in the open set experiments. They also considered verification using different distance measures and plotted the FAR against the FRR using DET curves. Overall, they found that color changing filters and, contrary to Botezatu \etal, that the filter that just occluded the mouth had little effect; however, filters that occluded the eyes had a significant negative effect. 

The work of Mirabet-Herranz \etal \cite{mirabet2022impact} is similar to the work of Hedman \etal in that they applied filters from Instagram to uncontrolled datasets of celebrity faces (in their case, the LFW and VIP\_Attribute datasets). Unlike the work of Hedman \etal that only chose Instagram filters that change the color tones of the image, Mirabet-Herranz \etal chose some Instagram filters that change the facial structure as well. Similarly to Hedman \etal they chose their filters based on popularity. (Given that both groups selected filters from Instagram based on popularity but resulted in different types of filters selected, we see that selecting filters based on popularity is a nonstandardized approach.) They tested their dataset on two gender classifiers and one weight estimator. In another work, Mirabet-Herranz \etal evaluated the impact of an additional six filters from Instagram, Snapchat, and TikTok on face verification, gender classification, apparent age estimation, weight estimation, and heart rate estimation, finding that most filters negatively impact soft biometric estimation, especially weight and heart rate networks, but some mild filters may enhance their performance~\cite{mirabet2024facial}.

Riccio \etal \cite{riccio2022openfilter} developed the OpenFilter framework to generate datasets of faces filtered with augmented reality filters from popular social media platforms using an Android Emulator. 
% We use their OpenFilter framework to assist in generating datasets of filtered faces.
% They also contributed a dataset of faces filtered with Instagram filters. 
They applied the filters to the FairFace \cite{karkkainen2021fairface} and Labeled Faces in the Wild \cite{huang2008labeled} datasets, and only used beautification filters (not fun/humorous filters). They selected the filters by popularity, which they assessed through articles in women's magazines. While this is a reasonable approach, popularity changes over time so it is not necessarily an enduring measure of the best filters. 
% The authors also studied two different research questions with their newly created datasets. 
They use the facial embeddings generated by the DeepFace \cite{taigman2014deepface}, VGGFace \cite{parkhi2015deep}, FaceNet \cite{schroff2015facenet}, CurricularFace \cite{huang2020curricularface}, MagFace \cite{meng2021magface}, and ElasticFace \cite{boutros2022elasticface} for their experiments.
% The first question is - Do beauty filters homogenize faces? They determined yes to this question. The second question is - Do beauty filters hinder facial recognition? This is a similar question to what we are asking in our study. They did this by comparing the difference in distance between filtered image pairs and original image pairs and perform paired t-tests of the similarity distributions of original faces and beautified faces.
They did not find that beautified faces had a big impact on facial recognition.

Tiwari \etal \cite{tiwari2023frll} presented the dataset FRLL-Beautified, which included faces filtered with filters from Snapchat, Faceapp, and B612. This study is the only one we have seen that uses filters from Snapchat on a controlled dataset \cite{debruine2017face}. They tested the different filters in their dataset by calculating the percentage of correct predictions. They only use 3 filters per app, which they hand-picked. They consider how filters affect accuracy of gender and ethnicity estimation, but not the effect on recognition.

Each of these previous works has demonstrated through various experiments and types of filters that overall, filters have a negative effect on facial recognition and other biometrics tasks. However, all of the previous works employ small numbers of handpicked filters. This paper builds on the foundation set by previous works, but our principled filter selection method allows for selection of a large and representative sample of filters from multiple social media platforms, enabling researchers to explore how the set of filters may affect recognition. Using our proposed framework, our case study builds on these previous works by using a greater number of filters and a greater diversity of filters than any of the previous works. Finally, an analysis on face embedding space reveals that, while some filters hurt recognition performance severely, this impact can be reduced significantly by using a simple linear transformation.

\section{The Framework}
This section introduces the components of our framework for studying how facial filters affect recognition on a controlled, large-scale, and per-app basis. Users of our framework must provide at least one facial filtering application and one facial recognition model.

\textbf{Component 1: Base Dataset of Controlled Face Images.}
We contribute a controlled and balanced dataset of face images for experimentation. This dataset enables researchers to perform experiments in a principled way without needing to generate their own dataset. Although the majority of filtered images today are taken with smartphone cameras, due to the ever-evolving computational photography techniques performed by smartphone cameras, in order to isolate the effects on recognition caused by a filter, as opposed to the camera itself, a controlled image set must be used. Thus, the images in our dataset were selected from the Facial Recognition Grand Challenge (FRGC)\cite{phillips2005overview} dataset: the standard dataset of controlled facial images across biometrics research. The FRGC dataset is available for free and is widely used in the biometrics research community. The filenames of our selected images are available in the supplementary material. The dataset contains 3,000 face images: 1,000 individuals with 3 different images per individual. These different images were taken on different days, with the subjects wearing different clothing, sometimes with different backgrounds. Our dataset is balanced by gender to support studies considering the effect of gender on recognition: 500 male and 500 female subjects are included. None of the 1,000 subjects have a twin among the other 999 subjects.

\textbf{Component 2: Filter Selection Method.}
The most important contribution of our framework is the filter selection process. Our dataset and experiments follow established practices in biometrics whereas the filter selection process is new. Our filter selection method moves away from hand-picking small numbers of filters based on a desired filter property to test (e.g., occlusions) or their popularity in a given moment. Instead, our method uses both qualitative (when possible) and quantitative means to select a large and representative number of filters from the chosen application. The qualitative means select for filters that make different types of modifications. The quantitative means select for filters that make different amounts of modifications. Together, the qualitative and quantitative means select a representative range of filters on the platform.

We propose the following filter selection method to select a representative range of filters from a particular application:
\begin{enumerate}
    \item Select a large number of filters, using qualitative means as much as possible to ensure diversity in this superset. These qualitative means include in-app filter categories that sort filters by type of modification.
    \item Apply each filter to the thirty total images of ten randomly-selected subjects from the dataset.
    \item Subtract the filtered images from the unfiltered images to find the mean difference per filter.
    \item Obtain binarized images using Otsu's method \cite{otsu1979threshold} to calculate the number of manipulated pixels.
    \item Divide the number of manipulated pixels by the size of the image to obtain the ratio of manipulated pixels.
    \item Group each filter into one of five bins by percentage of manipulated pixels: less than $20\%$, between $20\%$ and $40\%$, between $40\%$ and $60\%$, between $60\%$ and $80\%$ and greater than $80\%$.
    \item Select an equal number of filters per bin and apply these filters to all images in the dataset.
\end{enumerate}

This method selects filters with varying ranges of manipulated pixels in the image to ensure a mix filters that make small or large modifications are selected.

After the filters are selected, they can be applied to the base dataset using the OpenFilter \cite{riccio2022openfilter} framework.

\textbf{Component 3: Experiments.}
The final component of our framework is the experiments used for evaluating the filter's impact on recognition. We propose considering the genuine and impostor distributions for the original images as a baseline, then images with a particular filter compared to images with that same filter applied (\textit{Filtered v. Filtered}), and unfiltered images compared to images with a particular filter applied (\textit{Filtered v. Original}). We measure the distance between genuine and impostor distributions to assess the face recognition performance.

To calculate the genuine and impostor distributions for these different cases, one must first obtain facial embedding vectors of original and filtered faces using the desired facial recognition model. The protocol for performing one-to-one matching experiments on the filtered datasets is as follows. (The protocol for one-to-many experiments can be found in the supplementary material.) Cosine distance is used to measure similarity between two images. 

\textit{Original images.} Baseline results are obtained using the images without any filter. Genuine and impostor distributions are obtained for 1,000 subjects in our dataset. Since each subject has 3 images, 3,000 genuine and 4,495,500 impostor match scores are obtained for one-to-one analysis.

%For the one-to-one genuine distribution, 3000 similarity scores are generated from the 1000 subjects, 3 images per subject, and 3 different image pairs per subject. For the one-to-one impostor distribution, 3000 images will be compared with the 2997 (3000 total $-$ 3 of one subject) images of other subjects. Dividing by 2 to account for double-counted image pairs, this results in $(3* 1000 * 2997)/2 = 4,495,500$ similarity scores.

%For the one-to-many analysis, one of the three images of each person is set as the probe image and the other two images are used as gallery image. We select the maximum similarity score comparing each of the 1000 probe images to their corresponding 2 gallery images, for a total of 1000 mated scores. In the non-mated case, for each of the 1000 subjects, the probe image is compared with the all images of other subjects and the maximum score is picked to construct the non-mated distribution.

\textit{Filtered v. Filtered.}
Filtered versus filtered experiments are performed with a full set of images with the same filter applied, for every filter. In other words, images filtered with a particular filter are compared to images filtered with the same filter. Thus, the protocol and number of similarity scores are the same for the original v. original case.

\textit{Filtered v. Original.}
For the genuine case, each subject has images $A$, $B$, and $C$, and their filtered counterparts $A'$, $B'$, and $C'$. $A'$ is compared to $B$ and $C$, $B'$ is compared to $A$ and $C$, and $C'$ is compared to $A$ and $B$, resulting in 6 genuine scores. For the impostor case, filtered versions of each image are compared to the unfiltered versions of the other subjects, resulting in 3,000$\times$2,997 similarity scores.

\textit{Evaluation.}
Two measurements are used to quantify the effects of facial filters on face recognition systems. The distance between genuine and impostor distributions is measured using d-prime. Higher d-prime values mean better separation between these distributions, indicating better recognition accuracy. FNMR is used to report error rates.

% \vspace{-25pt}
\expandafter\def\expandafter\normalsize\expandafter{%
    \normalsize%
    \setlength\abovedisplayskip{0pt}%
    \setlength\belowdisplayskip{8pt}%
    \setlength\abovedisplayshortskip{-8pt}%
    \setlength\belowdisplayshortskip{2pt}%
}
\newcommand{\meanA}{\bar{X}_1}
\newcommand{\meanB}{\bar{X}_2}
\newcommand{\stdA}{\sigma_1}
\newcommand{\stdB}{\sigma_2}
% Calculate Cohen's d
\newcommand{\cohenD}{
  \frac{\meanA - \meanB}{\sqrt{\frac{\stdA^2 + \stdB^2}{2}}}
}
\begin{equation}
    d' = \cohenD
    \label{eq:dprime}
    \vspace{5pt}
\end{equation}

%(Unlike in the original v. original impostor experiment, in this case the similarity scores are not divided by 2. In the original v. original, image A is compared to image B and image B is compared to image A so the score is double-counted. However in this case image A' is compared to image B, and image B' is compared to image A, so there is no double counting.)

%We also separately track the genuine scores between pairs of images of the original and its filtered component, e.g. A' and A, and report these scores separately (see the supplementary material).
%One to many:
%In the mated case, for each of the 1000 subjects, we take the same image that was the probe in the original v. original and use the filtered version of that image as the probe. We use the other two original images as the gallery, so the same galleries as original v. original. Then we calculate the similarity scores between the probe and corresponding gallery of 2 images and take the max for all 1000 probes and their corresponding galleries, resulting in 1000 similarity scores.

%In the non-mated case, for each of the 1000 subject, we take the same image that was the probe in the original v. original and use the filtered version of that image as the probe. We use the 2997 unfiltered images of the other subjects as the gallery. Then we calculate the similarity scores between the 1000  probes and their corresponding galleries of 2997 images and take the max similarity score across all images in their galleries, resulting in 1000 similarity scores.

\section{Case Study}
We perform case studies of our framework to demonstrate its value.
For our case studies, we chose the American apps Instagram and Snapchat and the Chinese apps Meitu and Pitu as our filtering applications for cross-cultural analysis. We used the following pre-trained recognition models as our models \textit{m}: AdaFace (trained on WebFace12M \cite{zhu2021webface260m} dataset) \cite{kim2022adaface}, ArcFace (trained on MS1MV2 dataset) \cite{deng2019arcface}, FaceNet (trained on VGGFace2 dataset \cite{cao2018vggface2}) \cite{schroff2015facenet}, MagFace (trained on MS1MV2 dataset) \cite{meng2021magface} and VGGFace (trained on VGGFace dataset) \cite{parkhi2015deep}. ResNet-100 \cite{he2016deep} architecture is used for ArcFace, AdaFace and MagFace.

\textbf{Filter Selection.}
Here we discuss how we used each of the following filtering applications in our framework, and how the constraints of each application influenced the filter selection process. For Instagram, Pitu, and Meitu we used the Bluestacks Android emulator to run the apps within the OpenFilter framework. We were not able to run Snapchat on the Bluestacks emulator and instead ran it in the Microsoft app store. Across the four apps, we initially selected 326 filters. Following the filter selection process, we chose the total number of filters per app by selecting the bin with the smallest number of filters in it (excluding bins with 1 filter in them), and then selecting that number of filters from each of the other bins. Note that none of the four apps produced any filters in the last bin (more than $80\%$ manipulation). The number of initially selected filters is shown in Table \ref{tab:filter_bins}.

\begin{table}[tb]
% \vspace{-10pt}
% \vspace{-10pt}
\centering
\begin{center}
\caption{Number of initially selected filters in each bin. After the filter selection process, we select 32 filters from Instagram, 20 filters from Snapchat, 52 from Meitu and 21 from Pitu. The final number of selected filters are given in parentheses. Bin 1 has less than 20\% manipulation and Bin 5 has greater than 80\%.}
\label{tab:filter_bins}
\resizebox{\columnwidth}{!}{%
\begin{tabular}{c|c|c|c|c|c}
% \cline{2-6}
% \multicolumn{1}{c|}{} & \multicolumn{5}{c}{Number of filters } \\
% \cline{2-6}
\multicolumn{1}{c|}{} & Bin 1. & Bin 2. & Bin 3. & Bin 4. & Bin 5. \\
\cline{1-6}
Instagram & 49(8) & 18(8) & 25(8) & 8(8) & 0 \\
% \cline{1-6}
Snapchat & 13(10) & 10(10) & 1(0) & 1(0) & 0 \\
% \cline{1-6}
Meitu & 13(13) & 28(13) & 68(13) & 41(13) & 0 \\
% \cline{1-6}
Pitu & 34(7) & 1(0) & 7(7) & 9(7) & 0 \\
% \cline{1-6}
\end{tabular}
}
\end{center}
\vspace{-10pt}
\end{table}

\textit{Instagram.}
Instagram allows users to select filters from the following categories: Trending, Appearance, Aesthetic, Games, Humor and Special Effects. Filters in the Trending category include filters from the other categories, as well as filters that are not categorized. Users are also able to search keywords to find filters that are not present in any of these categories.

We initially chose 100 filters, aiming for a diverse range. Fifty-three filters are selected from the Trending (29), Appearance (8), Aesthetic (6), Humor (7) and Special Effects (3) categories. The remaining filters were selected using the following keywords: chin (1), mouth (10), eye (11), nose (4), beard (16) and glass (5). We applied each of these filters to 10 subjects' faces, and computed the difference between the original face and filtered face. We then binned the filters by the amount of difference between the original and the filtered face. We selected 8 filters per bin at random, for 32 filters total. Examples of binarized faces for each bin can be found in the supplemental material.

We then performed a post-hoc qualitative analysis of the selected filters to ensure our selection method maintained a diversity of filters. Of the 32 filters selected, 3 were categorized as Appearance, 3 were categorized as Aesthetic, and 1 was categorized as Special Effects. The majority of filters on Instagram are not categorized, so we speculatively categorized the other filters into these categories based on their visual properties. Our speculative categorizations for the remaining filters were 9 Appearance, 5 Aesthetic, and 11 Humor. The distribution of the selected filters across categories indicates that our filter selection method maintains a breadth of filter styles across chosen filters.

\textit{Snapchat.}
As mentioned earlier, we use the Snapchat application in Microsoft Store to generate filters, since we were not able to run the application using the Bluestacks Android Emulator. The version of Snapchat available on the Microsoft store only had 25 filters available. All available filters were applied to images of 10 subjects. After the binning process, 10 filters from the first and second bins were selected (the third and fourth bins had only 1 filter). 

\textit{Meitu.}
150 filters were initially picked randomly from all available 460 filters. The binning process resulted in the smallest bin having 13 filters in it, so 13 filters were selected from the first four bins, resulting in 52 filters total. These 52 filters were used to generate 156,000 filtered images.

\textit{Pitu.}
From all 56 filters, 51 filters were applied on 10 subjects (5 filters were ruled out because of similarity). After the binning process, 7 filters were selected from the first, third, and fourth bins. We used these 21 filters to generate 63,000 filtered images.

\begin{figure}[tb]
\centering
% original
\begin{subfigure}[t]{0.5\textwidth}
    \centering
    \begin{subfigure}[t]{0.15\textwidth}
        \centering
        \includegraphics[width=\linewidth]{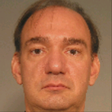}
    \end{subfigure}
            \caption{Original Image}
\end{subfigure}
\begin{subfigure}[t]{0.5\textwidth}
    \centering
    \begin{subfigure}[t]{0.15\textwidth}
        \centering
        \includegraphics[width=\linewidth]{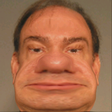}
    \end{subfigure}
    \begin{subfigure}[t]{0.15\textwidth}
        \centering
        \includegraphics[width=\linewidth]{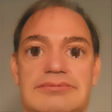}
    \end{subfigure}
    \begin{subfigure}[t]{0.15\textwidth}
        \centering
        \includegraphics[width=\linewidth]{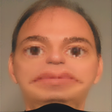}
    \end{subfigure}
    \begin{subfigure}[t]{0.15\textwidth}
        \centering
        \includegraphics[width=\linewidth]{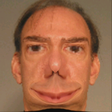}
    \end{subfigure}
    \begin{subfigure}[t]{0.15\textwidth}
        \centering
        \includegraphics[width=\linewidth]{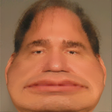}
    \end{subfigure}
        \caption{Instagram filters that have the highest impact on recognition}
\end{subfigure}

\begin{subfigure}[t]{0.5\textwidth}
\centering
    \begin{subfigure}[t]{0.15\textwidth}
        \centering
        \includegraphics[width=\linewidth]{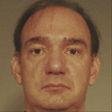}
    \end{subfigure}
    \begin{subfigure}[t]{0.15\textwidth}
        \centering
        \includegraphics[width=\linewidth]{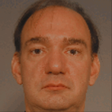}
    \end{subfigure}
    \begin{subfigure}[t]{0.15\textwidth}
        \centering
        \includegraphics[width=\linewidth]{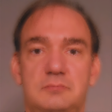}
    \end{subfigure}
    \begin{subfigure}[t]{0.15\textwidth}
        \centering
        \includegraphics[width=\linewidth]{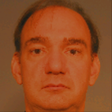}
    \end{subfigure}
    \begin{subfigure}[t]{0.15\textwidth}
        \centering
        \includegraphics[width=\linewidth]{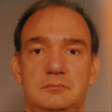}
    \end{subfigure}
    \caption{Instagram filters that have the lowest impact on recognition}
    \label{tab_best5_instagram}
\end{subfigure}

\centering
\begin{subfigure}[t]{0.5\textwidth}
    \centering
    \begin{subfigure}[t]{0.15\textwidth}
        \centering
        \includegraphics[width=\linewidth]{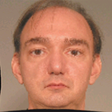}
    \end{subfigure}
    \begin{subfigure}[t]{0.15\textwidth}
        \centering
        \includegraphics[width=\linewidth]{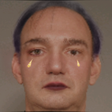}
    \end{subfigure}
    \begin{subfigure}[t]{0.15\textwidth}
        \centering
        \includegraphics[width=\linewidth]{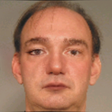}
    \end{subfigure}
    \begin{subfigure}[t]{0.15\textwidth}
        \centering
        \includegraphics[width=\linewidth]{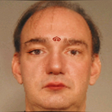}
    \end{subfigure}
    \begin{subfigure}[t]{0.15\textwidth}
        \centering
        \includegraphics[width=\linewidth]{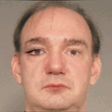}
    \end{subfigure}
        \caption{Meitu filters that have the highest impact on recognition}
\end{subfigure}

\begin{subfigure}[t]{0.5\textwidth}
\centering
    \begin{subfigure}[t]{0.15\textwidth}
        \centering
        \includegraphics[width=\linewidth]{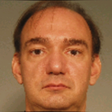}
    \end{subfigure}
    \begin{subfigure}[t]{0.15\textwidth}
        \centering
        \includegraphics[width=\linewidth]{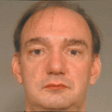}
    \end{subfigure}
    \begin{subfigure}[t]{0.15\textwidth}
        \centering
        \includegraphics[width=\linewidth]{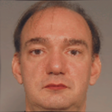}
    \end{subfigure}
    \begin{subfigure}[t]{0.15\textwidth}
        \centering
        \includegraphics[width=\linewidth]{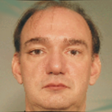}
    \end{subfigure}
    \begin{subfigure}[t]{0.15\textwidth}
        \centering
        \includegraphics[width=\linewidth]{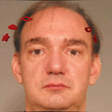}
    \end{subfigure}
    \caption{Meitu filters that have the lowest impact on recognition}

\end{subfigure}
\vspace{2pt}
    \caption{Example filters on Instagram and Meitu are shown. (a) 5 filters that have lowest d-prime values on Instagram ($4.59 \pm 0.51$). (b) 5 filters that have highest d-prime values on Instagram ($12.06 \pm 0.09$).
    (c) 5 filters that have the lowest d-prime values on Meitu ($8.69 \pm 0.23$). (d) 5 filters that have highest d-prime values on Meitu ($10.71 \pm 0.13$).}
    
    % Filters that change facial geometry cause the lowest d-prime values. The filters on Instagram tend to modify facial geometry more dramatically compared to Meitu filters.
    
    \label{fig:filters}
\vspace{-10pt}
\end{figure}

\begin{figure}[b]
    \centering
    \includegraphics[width=.5\textwidth]{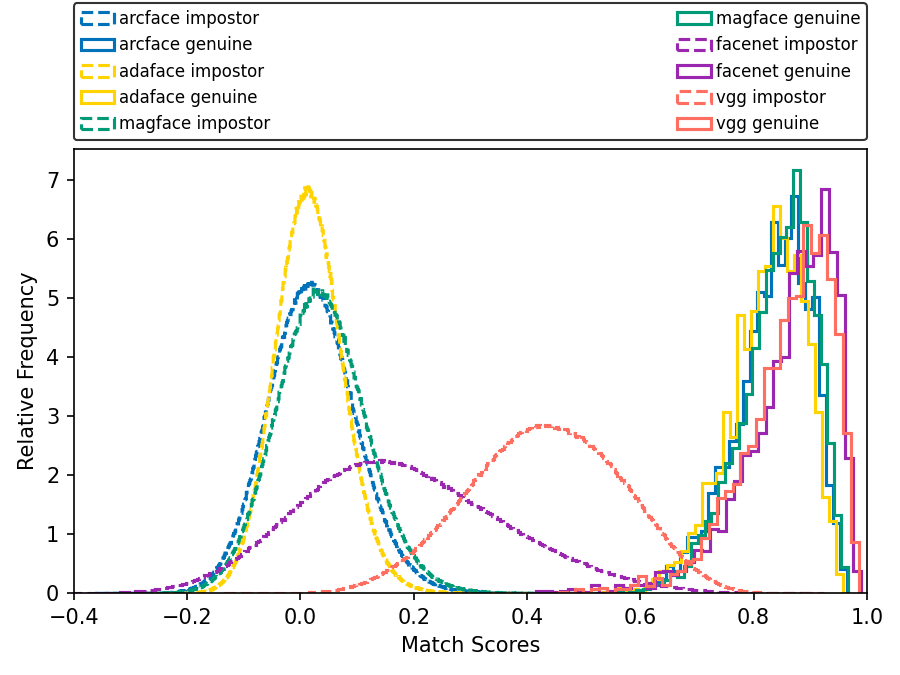}
    \caption{Genuine and impostor distributions for original images (without any filter). AdaFace model trained on WebFace12M achieves the highest d-prime among 5 models (12.66).}
    \label{fig:org}
\end{figure}

% \section{Case Study Results}

\textbf{Effect on Genuine and Impostor Distributions.}
For all filters and all matchers, we observed that the filtered v.~filtered experiments primarily shifted the impostor distributions and the filtered v.~original experiments primarily shifted the genuine distributions (the original v.~original genuine and impostor distributions can be seen in Figure \ref{fig:org}). The extent to which the genuine or impostor distribution shifted varied by filter. We observed similar results between Instagram and Snapchat and between Meitu and Pitu. Namely, we observed that Instagram and Snapchat had a wide variety of filters, some which shifted the genuine and impostor distributions greatly and others which shifted the genuine and impostor distributions minimally. In comparison, Meitu and Pitu filters had a milder effect on shifting the genuine and impostor distributions. Filtered v.~original distributions can be seen in Figures \ref{fig:instagram-original-filtered} and \ref{fig:meitu-original-filtered} for Instagram and Meitu (see supplementary material for the filtered v.~filtered experiment).

We find that filters that warp the facial geometry affect recognition the most, and filters that occlude the face also affect recognition. Filters that change the color  tones of the image do not affect recognition as much. A deeper analysis of the different types of filters that fall in each bin and their impact on recognition can be found in the supplementary material.

\begin{table}[t]
\centering
\begin{center}
\caption{d-prime values between genuine and impostor distributions for filtered v. original images. Results for male and female falls within the standard deviation range suggesting no major difference. While Meitu and Pitu filters do not have a significant variation, Instagram and Snapchat filters have greater range of impact.}
\label{tab:dprime_flt_v_org}
\resizebox{\columnwidth}{!}{%
\begin{tabular}{c|c|c|c|c|c}
% \cline{3-6}
\multicolumn{2}{c|}{} & Bin 1. & Bin 2. & Bin 3. & Bin 4. \\
\cline{1-6}
\multirow{3}{*}{Instagram} & all & $7.99 \pm 2.87$
& $7.43 \pm 2.27$ & $9.66 \pm 1.8$ & $11.37 \pm 1.38$ \\
& female & $7.72 \pm 2.75$ & $7.39 \pm 2.19$ & $9.41 \pm 1.7$ & $11.0 \pm 1.35$ \\
& male & $7.88 \pm 2.84$ & $7.25 \pm 2.26$ & $9.44 \pm 1.79$ & $11.16 \pm 1.29$ \\
\cline{1-6}
\multirow{3}{*}{Snapchat} & all & $8.09 \pm 2.65
$ & $5.15 \pm 3.14
$ & $-$ & $-$ \\
& female & $7.95 \pm 2.54
$ & $5.02 \pm 3.11
$ & $-$ & $-$ \\
& male & $7.95 \pm 2.62
$ & $5.13 \pm 3.09
$ & $-$ & $-$ \\
\cline{1-6}
\multirow{3}{*}{Meitu} & all & $9.75 \pm 0.59$ & $9.79 \pm 0.54$ & $9.57 \pm 0.59$ & $9.8 \pm 0.56$ \\
& female & $9.28 \pm 0.58$ & $9.35 \pm 0.55$ & $9.01 \pm 0.61$ & $9.36 \pm 0.68$ \\
& male & $9.79 \pm 0.61$ & $9.8 \pm 0.52$ & $9.77 \pm 0.55$ & $9.83 \pm 0.46$ \\
\cline{1-6}
\multirow{3}{*}{Pitu} & all & $10.09 \pm 0.35$ & $-$ & $10.05 \pm 0.22$ & $10.35 \pm 0.36$ \\
& female & $9.96 \pm 0.39$ & $-$ & $10.05 \pm 0.44$ & $10.26 \pm 0.43$ \\
& male & $9.77 \pm 0.44$ & $-$ & $9.65 \pm 0.34$ & $9.97 \pm 0.29$ \\
% \cline{1-6}
\end{tabular}
}
\end{center}
\vspace{-10pt}
\end{table}

% \begin{figure*}
%     \begin{subfigure}[b]{0.49\textwidth}
%         \includegraphics[width=\textwidth]{figures/instagram_aggregated_one-to-one-filtered-filtered-5best.png}
%         \label{fig:ig-filtered-filtered-5best}
%         \vspace{-10pt}
%         \caption{5 Best}
%     \end{subfigure}
%     %\vspace{-10pt}
%     \begin{subfigure}[b]{0.49\textwidth}
%         \includegraphics[width=\textwidth]{figures/instagram_aggregated_one-to-one-filtered-filtered-5worst.png}
%         \label{fig:ig-filtered-filtered-5worst}
%         \vspace{-10pt}
%         \caption{5 Worst}
%     \end{subfigure}
%     % \vspace{-10pt}
%     \caption{Aggregated graphs for the 5 best and 5 worst Instagram filters in terms of their impact on recognition for the filtered v. filtered experiments. We see that the 5 worst Instagram filters shift the impostor distribution towards the genuine.}
%     \label{fig:instagram-filtered-filtered}
% \end{figure*}

\begin{figure*}
    \begin{subfigure}[b]{0.49\textwidth}
        \includegraphics[width=\textwidth]{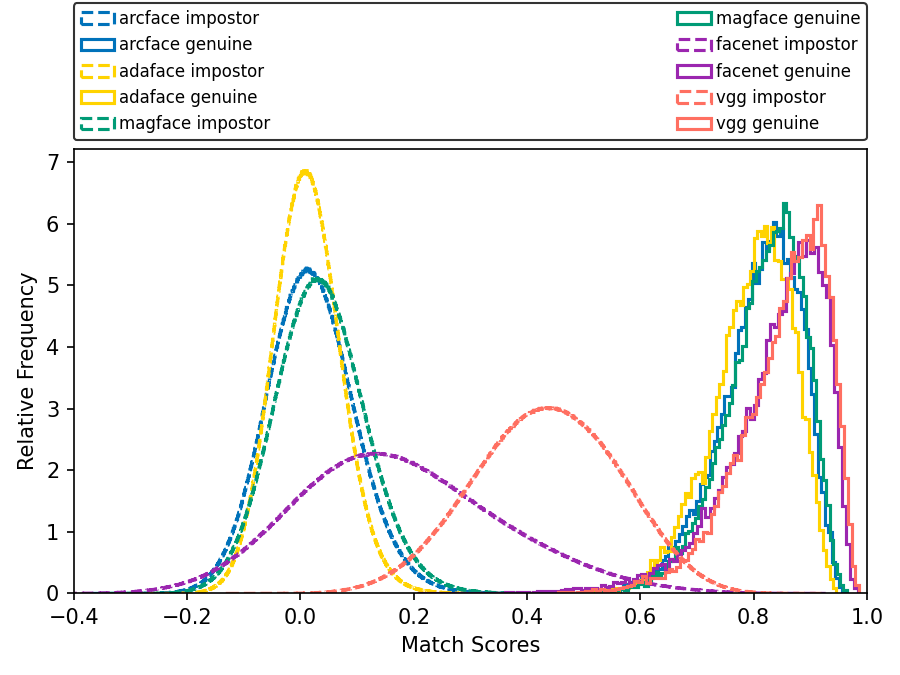}
        \label{fig:ig-original-filtered-5best}
        \vspace{-10pt}
        \caption{5 Best}
    \end{subfigure}
    %\vspace{-10pt}
    \begin{subfigure}[b]{0.49\textwidth}
        \includegraphics[width=\textwidth]{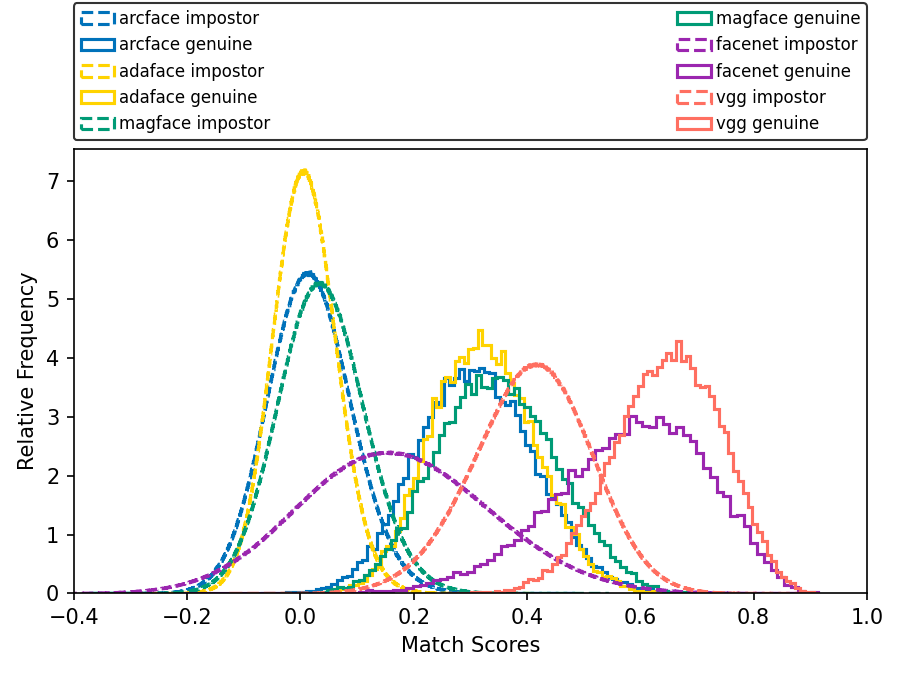}
        \label{fig:ig-original-filtered-5worst}
        \vspace{-10pt}
        \caption{5 Worst}
    \end{subfigure}
    % \vspace{-10pt}
    \caption{Aggregated graphs for the 5 best and 5 worst Instagram filters in terms of their impact on recognition for the filtered v.~original experiments. We see that the 5 worst Instagram filters shift the genuine distribution towards the impostor distribution.}
    \label{fig:instagram-original-filtered}
\end{figure*}

% \begin{figure*}
%     \begin{subfigure}[b]{0.49\textwidth}
%         \includegraphics[width=\textwidth]{figures/meitu_aggregated_one-to-one-filtered-filtered-5best.png}
%         \label{fig:meitu-filtered-filtered-5best}
%         \vspace{-10pt}
%         \caption{5 Best}
%     \end{subfigure}
%     %\vspace{-10pt}
%     \begin{subfigure}[b]{0.49\textwidth}
%         \includegraphics[width=\textwidth]{figures/meitu_aggregated_one-to-one-filtered-filtered-5worst.png}
%         \label{fig:meitu-filtered-filtered-5worst}
%         \vspace{-10pt}
%         \caption{5 Worst}
%     \end{subfigure}
%     % \vspace{-10pt}
%     \caption{Aggregated graphs for the 5 best and 5 worst Meitu filters in terms of their impact on recognition for the filtered v. filtered experiments. We see that the 5 worst Meitu filters slightly shift the impostor distribution towards the genuine, but not as dramatically as the 5 worst Instagram filters did.}
%     \label{fig:meitu-filtered-filtered}
% \end{figure*}

\begin{figure*}
    \begin{subfigure}[b]{0.49\textwidth}
        \includegraphics[width=\textwidth]{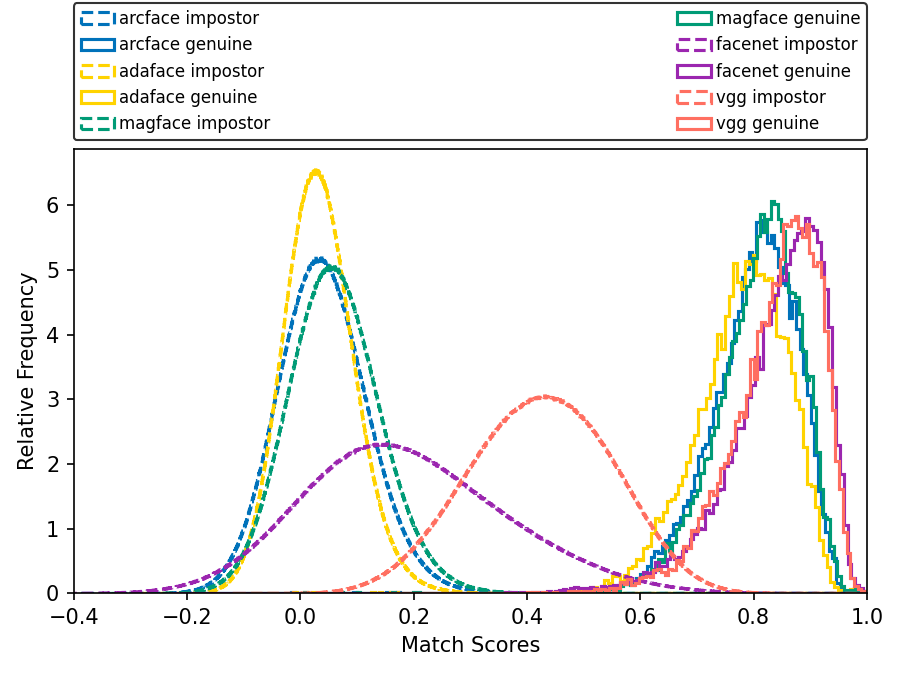}
        \label{fig:meitu-original-filtered-5best}
        \vspace{-10pt}
        \caption{5 Best}
    \end{subfigure}
    %\vspace{-10pt}
    \begin{subfigure}[b]{0.49\textwidth}
        \includegraphics[width=\textwidth]{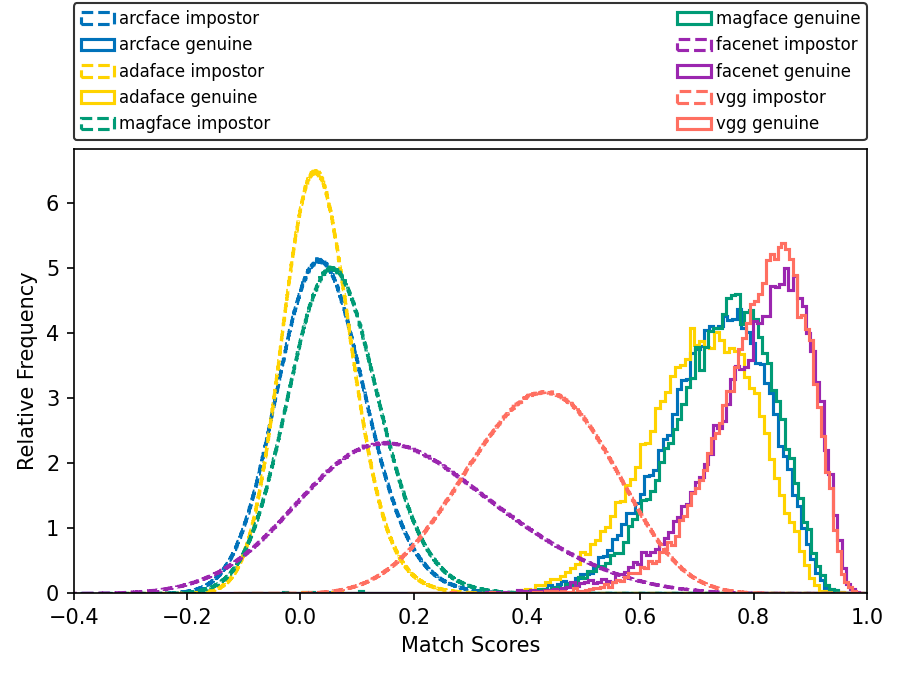}
        \label{fig:meitu-original-filtered-5worst}
        \vspace{-10pt}
        \caption{5 Worst}
    \end{subfigure}
    % \vspace{-10pt}
     \caption{Aggregated graphs for the 5 best and 5 worst Meitu filters in terms of their impact on recognition for the filtered v.~original experiments. We see that the 5 worst Meitu filters slightly shift the genuine distribution towards the impostor distribution, but not as dramatically as the 5 worst Instagram filters did.}
     \label{fig:meitu-original-filtered}
\end{figure*}

%Although different matchers produced different d-prime values based on the efficacy of that matcher, we observed that the different filters were generally consistent across matchers in how much they shifted the genuine or impostor distributions. Thus, for brevity, 
We report the d-prime between genuine and impostor distributions using the AdaFace model trained on the WebFace12M dataset, which had the best performance of our five models. 
%One-to-one analysis for original v. filtered (Table \ref{tab:dprime_flt_v_org}) and images is shown for 4 platforms (Instagram, Snapchat, Meitu and Pitu). 
Mean and standard deviation is reported per bin for all 1,000 subjects, 500 female subjects and 500 male subjects separately. Results for filtered v.~original analysis are given in Table \ref{tab:dprime_flt_v_org}. While the filtered v.~original and filtered v.~filtered experiments produce different genuine and impostor distributions, d-prime values between these distributions are close (see the supplementary material for filtered v.~filtered experiment) indicating similar recognition performance for filtered v.~original and filtered v.~filtered image pairs. Across bins, Snapchat and Instagram have greater variation in the d-prime, whereas Meitu and Pitu do not have much variation. We did not observe significant discrepancy between female and male subjects.

% \input{table_dprime_flt_v_flt}

%\textbf{Effect of Gender on Recognition.} For original v. original images we observe d-prime for males is slightly better than for females ($12.41$ against $12.19$). D-prime values between male and female subjects in original v. filtered and filtered v. filtered experiments fall within the standard deviation range, thus we cannot claim that there are filters that cause gender bias on recognition.

\section{Mitigating filter effect on face embeddings}

\begin{table*}[htb]
\centering
\begin{center}
% \normalsize
    \caption{Filter detection accuracy and FNMR results for the filtered v.~original experiment protocol (at 1-in-10,000 and 1-in-100,000 FMR) are given for 4 platforms. Mean and standard deviations are reported for 5 filters that has the highest impact on recognition. Proposed filter effect mitigation approach ($FNMR_{mapping}$) reduces error rates significantly on Instagram and Snapchat filters over the pretrained baseline ($FNMR_{pre}$).}
    \label{tab:fnmr}
    \centering
    
    \begin{tabular}{c||c||cc|cc}

      & \makecell{Filter \\ Detection} $\uparrow$ & \makecell{$FNMR_{pre}$ \\ $(10^{-4} FMR)$} $\downarrow$ & \makecell{$FNMR_{mapping}$ \\ $(10^{-4} FMR)$} $\downarrow$ & \makecell{$FNMR_{pre}$ \\ $(10^{-5} FMR)$} $\downarrow$ & \makecell{$FNMR_{mapping}$ \\ $(10^{-5} FMR)$} $\downarrow$ \\
      \hline
    
    Instagram & $100.00 \pm 0.00\%$ & $0.217 \pm 0.115\%$ & $0.015 \pm 0.017\%$ & $0.393 \pm 0.186\%$ & $0.033 \pm 0.034\%$ \\
    \hline
    
    Snapchat & $100.00 \pm 0.00\%$ & $0.756 \pm 0.234\%$ & $0.391 \pm 0.350\%$ & $0.850 \pm 0.176\%$ & $0.496 \pm 0.354\%$ \\
    \hline
    
    Meitu & $88.81 \pm 0.71\%$ & $0.003 \pm 0.002\%$ & $0.003 \pm 0.002\%$ & $0.008 \pm 0.023\%$ & $0.007 \pm 0.024\%$ \\
    \hline
    
    Pitu & $82.92 \pm 0.85\%$ & $0.003 \pm 0.002\%$ & $0.003 \pm 0.002\%$ & $0.024 \pm 0.103\%$ & $0.024 \pm 0.102\%$ \\
    \end{tabular}
\end{center}
    
\end{table*}

Beyond the face similarity analysis presented in the previous section, we further analyze the impact of the filtered face images on the embedding space of the pretrained face recognition model. Two models on top of face embeddings are used to mitigate filter effect. First, we train a classifier to detect the face filters to see whether the embedding space carries sufficient information to classify filters. Then, given the detected filters we train a linear layer to map representations of filtered images to original images to see if transformed representations help alleviate the performance drop on recognition caused by face filters.

We use subject-disjoint splits for our experiments: 700 subjects for training, 100 for validation, and 200 for testing. The experiments were repeated with five randomly selected data splits. Images from the training set are used to train both the filter classification model and the embedding transformation model. Both models only include a single linear layer that takes a 512-D embedding vector as input. For filter classification, 5 filters from each platform that have the highest negative impact on recognition are used (20 filters). With the original images the model is trained to recognize 21 classes using the Adam optimizer \cite{adam} minimizing the cross-entropy loss. Training is stopped if the validation loss did not improve for 50 consecutive epochs. Next, we trained a second model to map filtered images to original images. A single linear mapping is trained separately for each filter to minimize mean squared error (MSE) between embeddings of the original and filtered images. Again, we used the same early stopping criteria to terminate training. This resulted in 512-D mapped representations to mitigate the effects of face filters.

\begin{figure}[t]
    \centering
    \includegraphics[width=.5\textwidth]{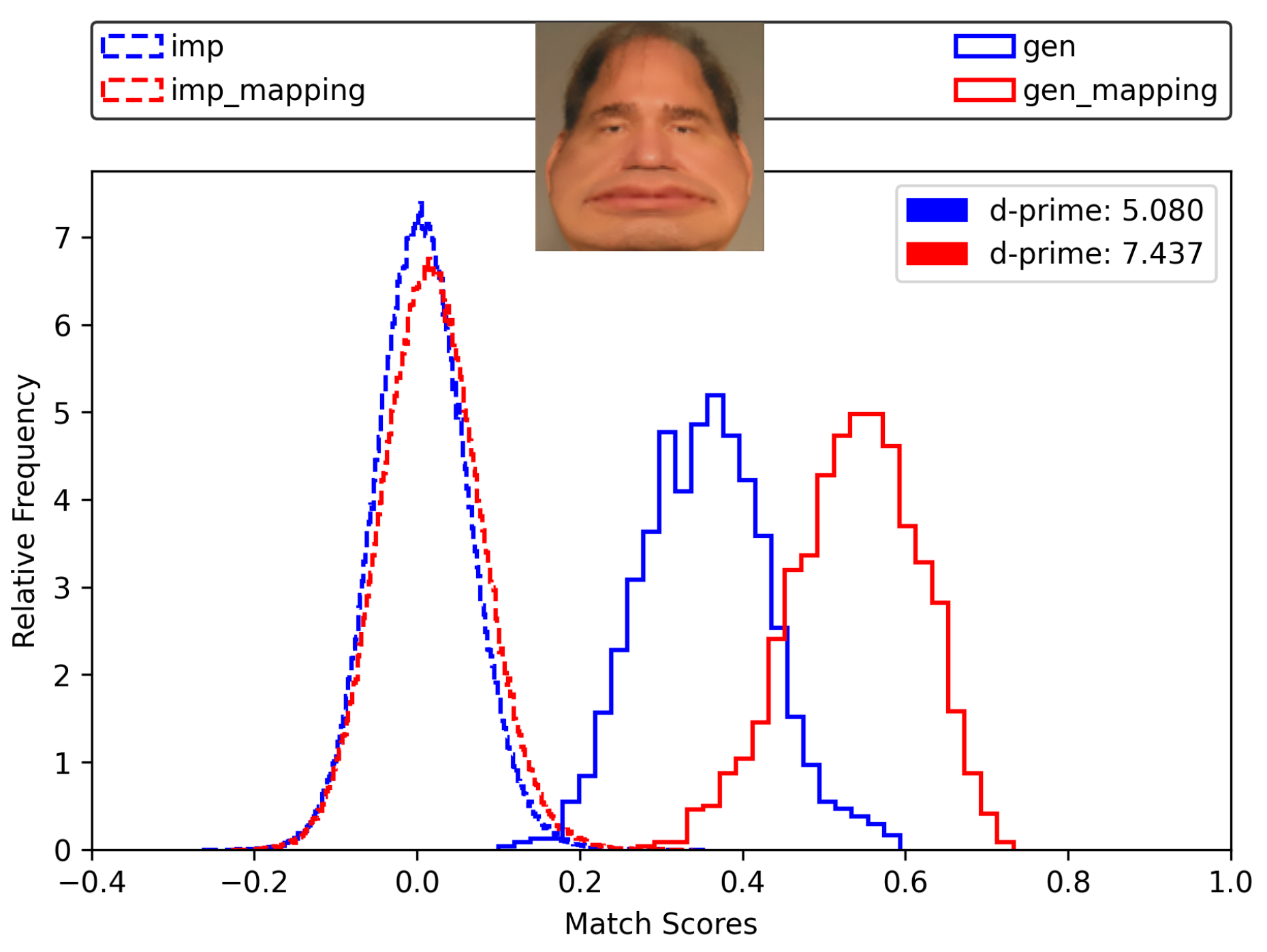}
    \caption{Filtered v.~original similarity score distributions are given for the depicted face filter. The d-prime values indicate improved separability between impostor and genuine scores using the proposed approach, shifting genuine distribution to higher scores and significantly reducing FNMR, as given in Table \ref{tab:fnmr}.}
    \label{fig:finetune}
\end{figure}

Filter classification accuracy and FNMRs at two different thresholds are given in Table \ref{tab:fnmr} for five filters that have the highest impact on recognition. Filter detection results show that filters that cause greater increase in error rates are easier to detect, resulting in perfect accuracy on Instagram and Snapchat filters. Additionally, for these filters FNMRs are significantly improved with the proposed linear representation transformation. Results indicate that, while face filters can cause crucial accuracy drop on recognition, this impact can be mitigated by learning a linear transformation.

\section{Discussion}

Our framework identifies a large and representative set of filters for each of our four platforms, and through the presented the case study, we uncover general trends about how each platforms' filters affect recognition performance. Analysis on face embeddings show that, while accuracy drop can be significant for some filters, this impact can be restored in the latent space efficiently to enhance recognition performance. Similarity score distributions for an example filter, using a pretrained recognition model (depicted with blue) and the proposed embedding transformation (depicted with red) can be found in Figure \ref{fig:finetune}.

\textbf{Prevalence and Impact of Facial Geometry-Modifying Filters.}
We observed that the impact of a filter on face recognition is mostly dependent on how much the filter morphs the facial geometry in the image. We see in Figure \ref{fig:filters} that the 5 worst Instagram images all morph the facial geometry drastically, whereas the 5 best filters may change the color tones of the image but don't impact the face's geometry. We only had one Meitu filter that significantly changed facial geometry, leading to the most severe degradation in recognition performance, while the remaining filters had a more moderate impact on accuracy. The color-changing filters' limited effect on recognition is consistent with previous literature indicating that deep CNN face matchers perform similarly on grayscale and colored images \cite{bhatta2024our}.

Previous work studying the impact of facial filtering on recognition has primarily considered filters that occlude the face or change the color tones of the image. Our case study uncovers that especially on Western social media applications, filters that modify facial geometry are quite prevalent. This points to the importance of future work that more deeply studies the effects of changes to facial landmarks on recognition; for example, studying which changes to facial landmarks have the greatest impact on recognition, or what types of changes to facial landmarks are most prevalent in current social media filters.

\textbf{Comparison of American and Chinese Filters.}
Our filter selection process and experimental results can also shed light on the social implications of facial filtering from a cross-cultural perspective. Previous work by Conwill \etal~\cite{conwill2024virtualization} on the social impacts of facial filtering has concluded that filters that confound human facial perception are more likely to have a negative impact on self-esteem, whereas filters that cause the face to be perceived differently from the original may promote imagination and creativity. Although human facial recognition and automated facial recognition are not exactly the same, a filter's impact on automated facial recognition can be an indication as to if that filter will confound facial perception as well. This claim is further supported by the fact that the impact of a filter on face recognition is mostly dependent on how much the filter morphs the facial geometry in the image. The 5 worst Instagram filters all morph the facial geometry drastically, whereas the 5 best filters may change the color tones of the image but don't impact the face's geometry much. The same trend occurs with the Snapchat filters. Only one Meitu filter significantly changes facial geometry, and it had the worst impact on recognition of all the Meitu filters. 

%The filters that modify facial geometry are more similar to the ``creative'' filters discussed in Conwill \etal, whereas filters that don't modify facial geometry tend to be more similar to the ``traditional'' filters~\cite{conwill2024virtualization}.

The American apps Instagram and Snapchat had a wide range of filters, some of which did not shift the genuine or impostor distributions much and others which shifted them a lot. On the other hand, the Chinese apps Meitu and Pitu only had filters that shifted the distributions slightly. While Snapchat and Instagram both have filters that modify the facial structure minimally and filters that modify the facial structure a lot, most of the Meitu filters just add makeup or change the color tones of the image slightly, and a few change the facial structure. The Pitu filters only add makeup or change the color tones of the image. The employed filter selection process attempts a representative sampling of the available filters, meaning that the Meitu and Pitu filters included in the study were not incidentally just light filters; rather, these light filters are a representative sampling of the filters available on these apps.

One possible reason for this difference is that Asian and Western cultures have different motivations for filter usage. One study found both that Asians experience more external societal pressure to look pretty and to display themselves through a beauty filter, and also that Americans tend to prefer humorous filters while East Asians prefer more natural-looking beauty filters \cite{herringstrategic}. In China, beauty filtering is even associated with female empowerment and inclusion in professional spaces \cite{peng2021techno}. We found that changes in facial geometry are the primary factor indicating how a filter will affect recognition, and the humorous filters found in American apps affect recognition more than any minor changes in facial geometry (if at all) that the Chinese apps perform.

\section{Conclusion}

In this work, we propose a framework to investigate the effect of facial filters on face recognition. While previous works analyze this phenomenon, they only use few hand-picked filters for their evaluation. Our work presents a filter selection methodology to ensure collection of a large and diverse set of filters for analysis on a controlled dataset. 

We demonstrate our framework with a case study consisting of 125 facial filters from 4 different social media applications. We observe that filters in the Chinese applications make subtle modifications on a face for aesthetic concerns, while Instagram and Snapchat have a number of filters that cause large facial deformation for humorous purposes: this preference for humor in American culture over aesthetics in Chinese culture can result in a significant decrease in recognition performance. The analysis on the latent space reveals that the accuracy drop caused by these filters can be mitigated by restoring the face embedding with a linear mapping. Our work, facilitated by a large and diverse set of filters, advances the understanding of how facial filters influence face recognition systems and offers an effective method to preserve high accuracy rates even in the presence of heavily modified images.

{\small
\bibliographystyle{ieee}
\bibliography{egpaper_for_review}
}

\end{document}